\documentclass[11pt]{article}

\usepackage[preprint]{acl}

 \usepackage{microtype}

\usepackage{hyperref}       
\usepackage{url}            
\usepackage{booktabs}       
\usepackage{amsfonts, bm}       
\usepackage[mathscr]{euscript}
\usepackage{nicefrac}       
\usepackage{microtype}      
\usepackage{xcolor}         
\usepackage[normalem]{ulem}
\usepackage{soul}
\usepackage{enumitem}
\usepackage{bbm}
\usepackage{subfig}
\usepackage{array}          
\usepackage{multirow}
\usepackage{amsmath}
\usepackage{scalerel,graphicx,xparse}

\usepackage[textwidth=2cm]{todonotes}
\usepackage{listings}
\usepackage{xcolor}
\usepackage{hwemoji}
\usepackage{fontawesome5}
\usepackage{xcolor} 

\newcommand{\frozen}{\faSnowflake}
\NewDocumentCommand\trainableNew{}{\scalerel*{\includegraphics{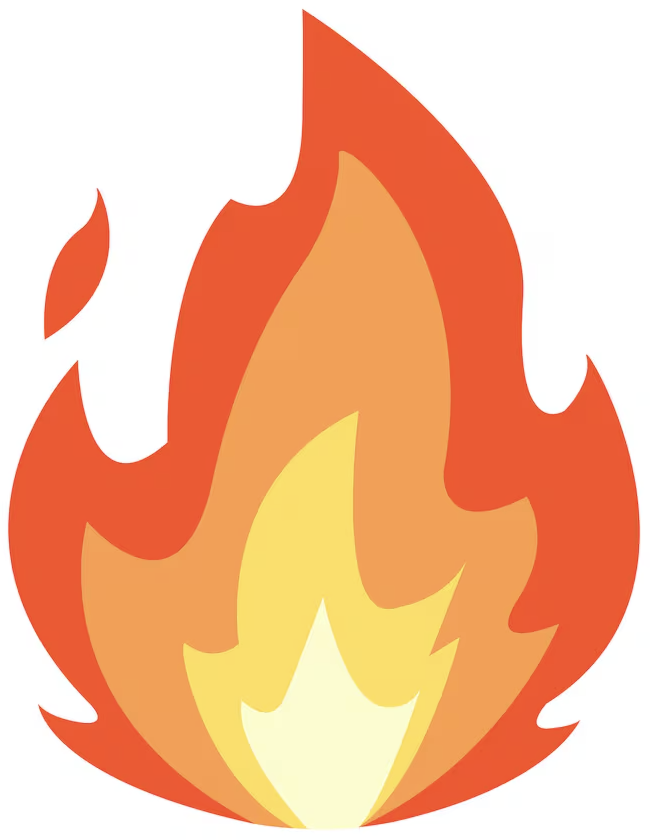}}{X}}
\usepackage{cleveref}

\newif\ifdraft
\drafttrue

\ifdraft
    
    \newcommand{\TODO}[1]{\todo[size=\tiny]{#1}}
    \newcommand{\todoInline}[1]{{\color{red}{\bf[TODO: #1]}}} 
    \newcommand{\highlight}[1]{{\color{blue}#1}}
    \newcommand{\pubapprove}[1]{{\color{olive}#1}}
    \newcommand{\nilesh}[1]{\todo[color=cyan,size=\tiny]{NG: #1}}
    \newcommand{\pj}[1]{\todo[color=yellow,size=\tiny]{PJ: #1}}
    \newcommand{\ak}[1]{\todo[color=orange,size=\tiny]{AK: #1}}
    \newcommand{\devvrit}[1]{\todo[color=red,size=\tiny]{DK: #1}}
    
\else 
    
    \newcommand{\TODO}[1]{}
    \newcommand{\todoInline}[1]{}
    \newcommand{\highlight}[1]{}
    \newcommand{\pubapprove}[1]{}
    \newcommand{\pj}[1]{}
    \newcommand{\ak}[1]{}
    \newcommand{\nilesh}[1]{}
    \newcommand{\devvrit}[1]{}
    
\fi

\title{Compressing Many-Shots in In-Context Learning}

\author{
  Devvrit Khatri\thanks{\hspace{1.5mm}Correspondence to: devvrit.03@gmail.com, prajain@google.com}$^{\; \delta\diamond}$ \enspace
  Pranamya Kulkarni$^{\diamond}$ \enspace
  Nilesh Gupta$^{\delta\diamond}$ \enspace
  Yerram Varun$^{\diamond}$ \\[1.5mm] 
  \textbf{Liqian Peng$^{\diamond}$ \enspace
  Jay Yagnik$^{\diamond}$ \enspace
  Praneeth Netrapalli$^{\diamond}$ \enspace
  Cho-Jui Hsieh$^{\dagger\diamond}$} \\[1.5mm] 
  \textbf{Alec Go$^\diamond$ \enspace
  Inderjit S Dhillon$^{\delta \diamond}$ \enspace
  Aditya Kusupati$^{\diamond}$ \enspace
  Prateek Jain\footnotemark[1]$^{\; \diamond}$} \\[2mm] 
  $^\diamond$Google \enspace $^\dagger$UCLA \enspace $^\delta$University of Texas at Austin
}

\begin{document}

\maketitle

\begin{abstract}
Large Language Models (LLMs) have been shown to be able to learn different tasks without explicit finetuning when given many input-output examples / demonstrations through In-Context Learning (ICL). Increasing the number of examples, called “shots”, improves downstream task performance but incurs higher memory and computational costs. In this work, we study an approach to improve the memory and computational efficiency of ICL inference by compressing the many-shot prompts. Given many shots comprising $t$ tokens, our goal is to generate a $m$ soft-token summary, where $m < t$. We first show that existing prompt compression methods are ineffective for many-shot compression, and simply using fewer shots as a baseline is surprisingly strong. To achieve effective compression, we find that: (a) a stronger compressor model with more trainable parameters is necessary, and (b) compressing many-shot representations at each transformer layer enables more fine-grained compression by providing each layer with its own compressed representation. Based on these insights, we propose \textbf{MemCom}, a layer-wise compression method.
We systematically evaluate various compressor models and training approaches across different model sizes (2B and 7B), architectures (Gemma and Mistral), many-shot sequence lengths (3k-6k tokens), and compression ratios ($3\times$ to $8\times$).
MemCom outperforms strong baselines across all compression ratios on multiple classification tasks with large label sets. 
Notably, while baseline performance degrades sharply at higher compression ratios, often by over 20-30\%, MemCom maintains high accuracy with minimal degradation, typically dropping by less than 10\%.
\end{abstract}
\section{Introduction}
In-Context Learning  (ICL)~\citep{icl_brown_2020} is a method by which language models learn a task by conditioning on examples of that task, provided as part of the input. Large Language Models (LLMs) have demonstrated remarkable abilities to generalize to new tasks simply by attending to a sequence of input-output pairs -- commonly referred to as ``shots" -- followed by a test input. This approach has gained widespread popularity due to its flexibility and minimal setup. Unlike finetuning, ICL requires no updates to model parameters. As a result, the same model can be used for all downstream tasks without any finetuning overhead.

A common assumption has been that finetuning typically yields superior performance on downstream tasks compared to ICL. However, recent studies~\citep{agarwal2024many,bertsch2024context} demonstrate that this performance gap narrows significantly as the number of in-context examples increases. As the support for long context for LLMs increases, many-shot ICL poses as an attractive and practical alternative to finetuning -- retaining the benefits of task flexibility and reduced overhead, while delivering improved accuracy through richer context.
\begin{figure*}[!th]
\centering
\includegraphics[width=.9\textwidth]{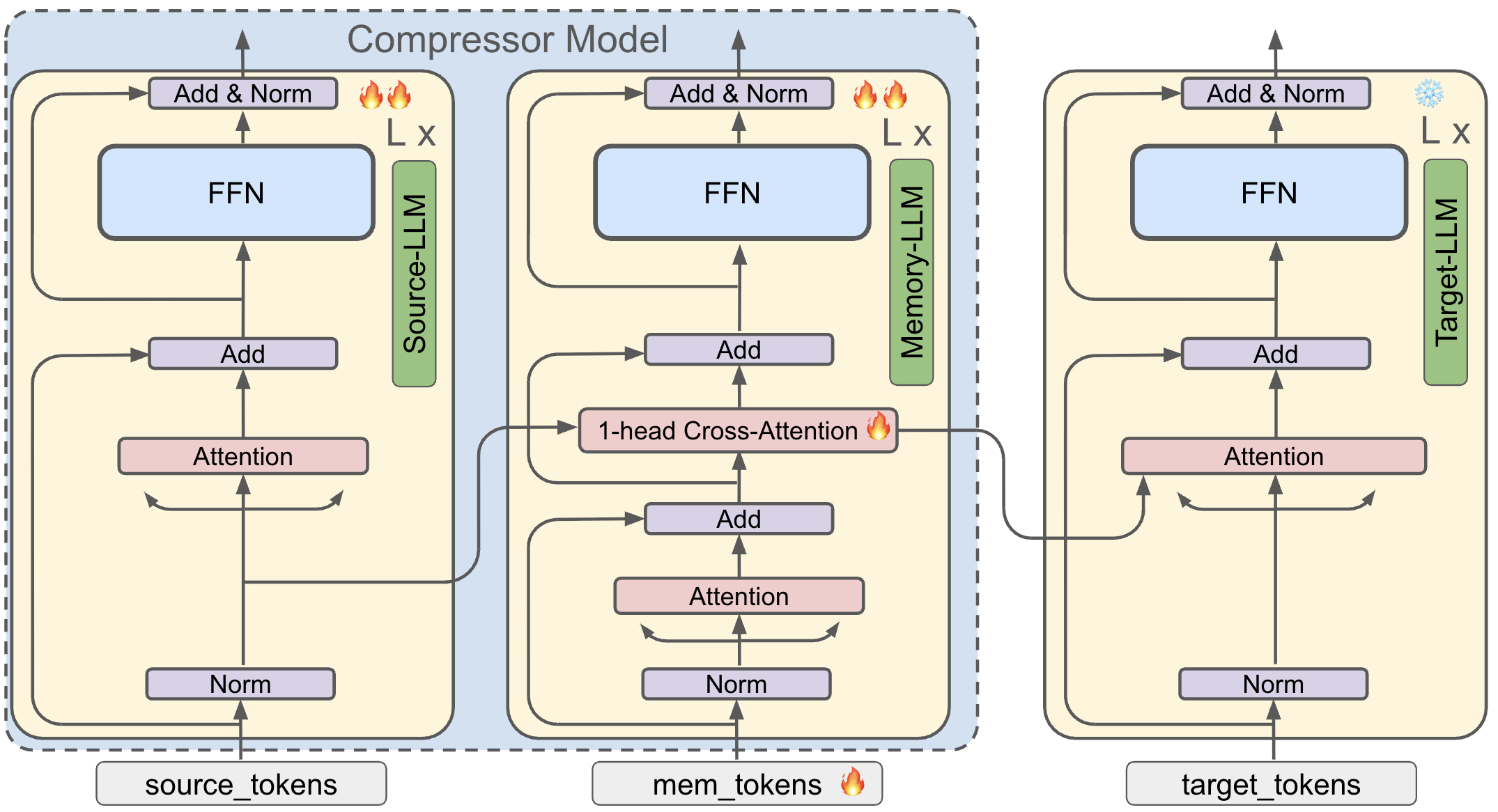}
\caption{ \small MemCom consists of two stacks of LLMs working as compressor. The Source-LLM processes the input tokens to be compressed. The Memory-LLM processed memory tokens, which cross-attends to the input representation in corresponding layer. Finally, during inference, Target-LLM (frozen \frozen) processes test input and attends to the compressed representations at each layer. Training happens in two phases. In Phase-1, only the cross-attention module and the memory tokens are trained (represented by \trainableNew), while in Phase-2 the entire stack of Source-LLM and Memory-LLM is trained (\trainableNew\trainableNew), in addition to the memory tokens.} 

\label{fig:memcom}
\end{figure*}



Despite its promise, many-shot ICL introduces its own set of challenges, primarily in terms of memory and computation overhead during inference. To support many-shot learning, the key-value (KV) cache corresponding to all example tokens must be stored and maintained, leading to significant memory usage. Additionally, each test input must attend to these cached examples, which results in substantial computational cost—especially as the number of shots increases. This creates a bottleneck that limits the scalability of many-shot ICL in practical deployments.

In this work, we address the above challenge by introducing a LLM-based compression technique, MemCom, to form compressed KV cache. At inference time, the model attends only to this compressed cache, significantly reducing memory usage and attention computation. A practical use case for this approach arises in edge deployments of LLMs, where resource constraints demand efficient inference. In such scenarios, a hybrid Cloud-Edge architecture can be employed: the cloud performs the many-shot compression offline, and the resulting compressed representation is used by the edge device during inference. Hybrid strategies leveraging cloud-edge collaboration have been explored in various domains~\citep{Almeida_2022,chen2024netgpt,Laskaridis_2020}, including recent work that customizes edge LLMs using LoRA adapters hosted in the cloud~\citep{bang2024crayon}. Similarly, our approach enables offline compression of in-context examples, making many-shot ICL practical and efficient for deployment in constrained environments.
\begin{figure*}[th!]
    \centering
    \subfloat[]{\includegraphics[width=0.24\textwidth]{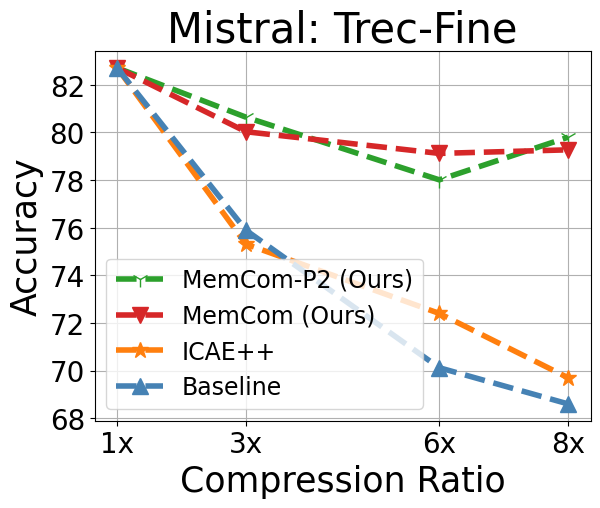}}
    \hfill
    \subfloat[]{\includegraphics[width=0.24\textwidth]{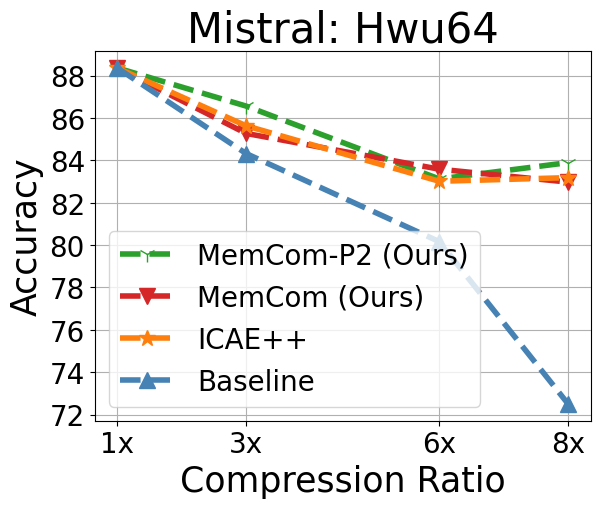}}
    \hfill
    \subfloat[]{\includegraphics[width=0.24\textwidth]{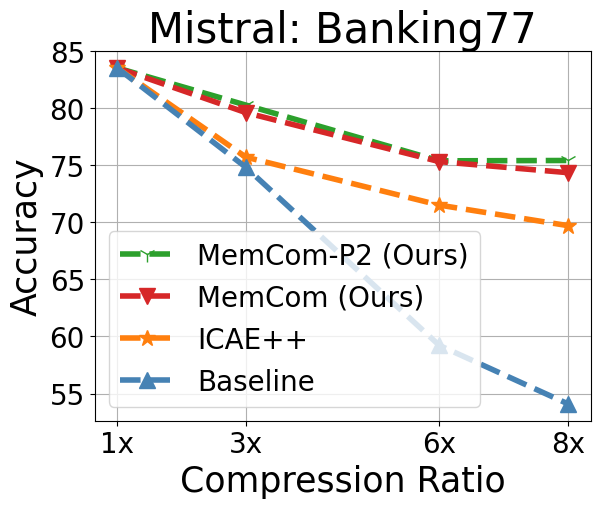}}
    \hfill
    \subfloat[]{\includegraphics[width=0.25\textwidth]{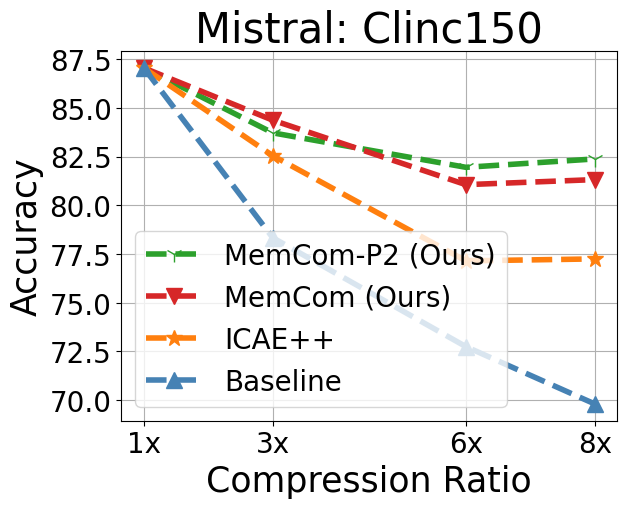}}\\
    \subfloat[]{\includegraphics[width=0.24\textwidth]{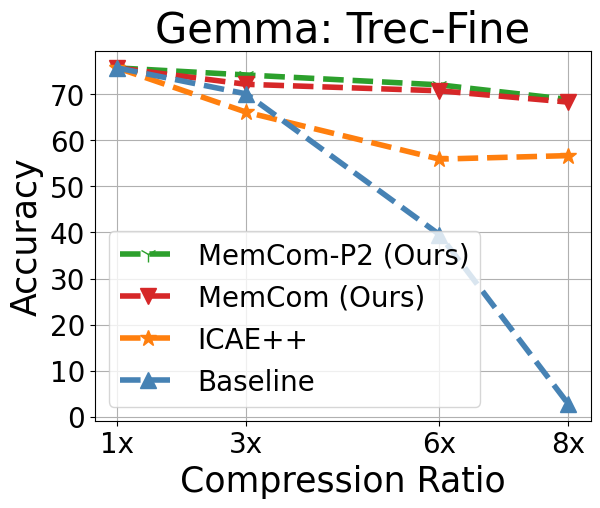}}
    \hfill
    \subfloat[]{\includegraphics[width=0.24\textwidth]{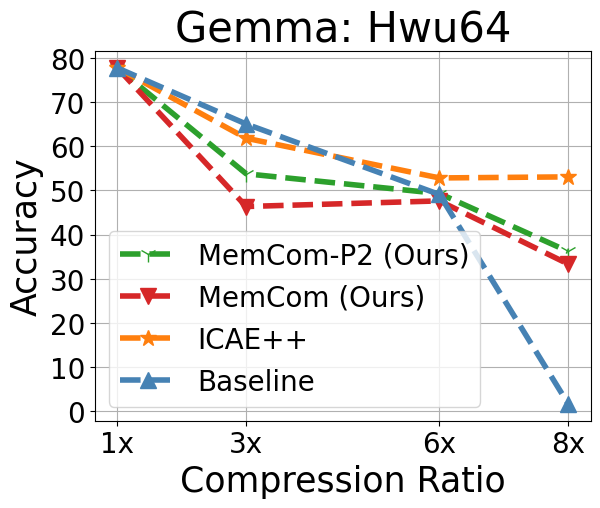}}
    \hfill
    \subfloat[]{\includegraphics[width=0.24\textwidth]{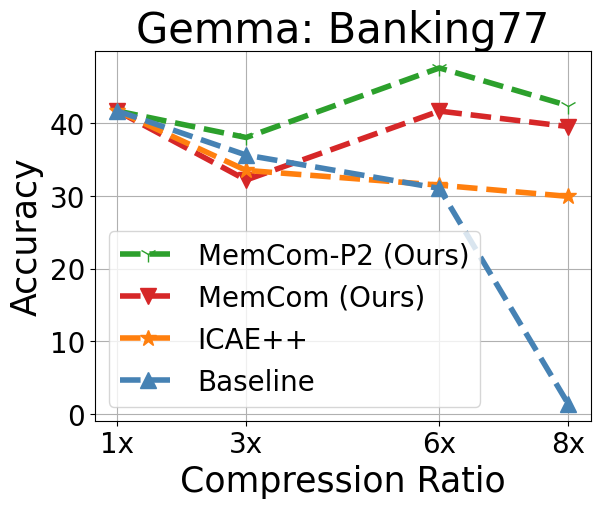}}
    \hfill
    \subfloat[]{\includegraphics[width=0.24\textwidth]{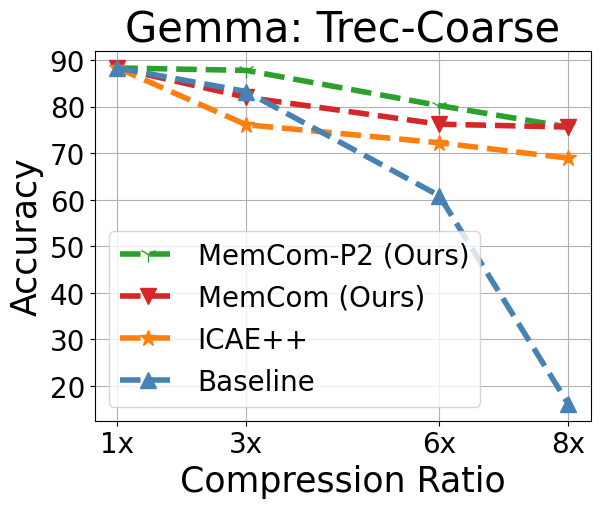}}
    \caption{Comparison of compression performance across different compression ratios. As the compression ratio increases, the vanilla baseline (``Baseline") exhibits significant performance degradation. ICAE++ performs better than the vanilla baseline but remains less robust compared to MemCom, which maintains high performance even at high compression ratios. The top row shows results for Mistral-7B, where sequences of $6k$ tokens are compressed. The bottom row shows results for Gemma2-2B, with $3k$-token sequences being compressed. MemCom refers to our method's performance post Phase-1 training, which introduces comparatively lightweight compressor of 1-head cross attention in every layer. While MemCom-P2 refers to the performance post Phase-2 training, which has two full stacks of LLMs working as the compressor. Section~\ref{sec:memcom_arch} introduces the training methodology.}
    \label{fig:compression_fig}
\end{figure*}

MemCom consists of two stacks of LLMs working as compressor (Figure~\ref{fig:memcom})
producing a compressed KV cache of the many-shot input across layers. The Source-LLM (left) processes the original many-shot $t$ input tokens. The Compressor-LLM (middle) takes a fixed set of $m$ memory tokens and, at each layer $i$, performs cross-attention over the $i$-th layer representations of the $t$ input tokens in the Source-LLM. The resulting compressed memory representations are passed to the Target-LLM (right), which is the deployed model responsible for inference. It processes the test inputs and, at each layer, attends to the compressed representations instead of the full input. Note that the target-LLM is frozen, and not finetuned.
Since compression is performed offline, we opt for a powerful compressor to ensure high-quality, near-lossless compression. At inference time, the model only needs to attend to $m$ tokens rather than the original $t$, substantially reducing memory and compute.

To the best of our knowledge, this is the first work to explore many-shot compression for in-context learning. While several prompt compression techniques have been proposed ~\citep{mu2023learning,ge2024context,chevalier2023adapting}, they are primarily designed for short prompts (typically <1k tokens) and often operate by compressing only the final hidden states. We tested the most relevant recent method, ICAE~\citep{ge2024context}, to the demanding long-context many-shot setting, and found that it does not retain performance. Our analysis suggests two primary bottlenecks with such approaches for many-shot compression: first, the limited capacity of the compressor model, and second, the coarse nature of compressing using only final-layer compressed representations. Interestingly, a simple baseline of using fewer shots to fit the m-token budget proves surprisingly strong. In addition to increasing compressor capacity (i.e., using more trainable parameters), we identify that a more fine-grained compression strategy is crucial for improving compression quality. Thus, MemCom introduces a strong layer-wise compression, a key departure from prior methods, enabling each transformer layer to access its own tailored compressed representation of the many-shot context.

We perform evaluations on five classification tasks with large
label sets (Table~\ref{tab:datasets}), and demonstrate that MemCom consistently outperforms the strongest baseline across compression ratios ranging from $3\times$ to $8\times$
Notably, as the compression ratio increases, the performance of the few-shot baseline degrades sharply, whereas MemCom maintains robust accuracy with only marginal decline (Figure~\ref{fig:compression_fig}).
\section{Related Work}
\paragraph{The impact of scaling many-shots in In-Context Learning.}
\citet{brown2020language} demonstrated that scaling LLMs significantly enhances their few-shot learning capabilities, paving the way for models like GPT-3. Subsequent work~\citep{agarwal2024many,bertsch2024context,li2023context} explored the benefits of many-shot ICL, showing that increasing the number of demonstrations leads to substantial performance gains, often rivaling supervised fine-tuning. This underscores the potential of scaling demonstrations in ICL while emphasizing the growing need for efficient methods to manage the associated computational costs.
\paragraph{Shot-selection and KV cache compression.}
Previous works have explored demonstration selection as a strategy to improve ICL efficiency~\citep{dong-etal-2024-survey,luo2024iclretrieval}. However, these approaches still require storing the full set of many-shot examples, leaving the memory overhead issue unaddressed. Additionally, they operate entirely at inference time by dynamically retrieving relevant examples based on the test query.  
In contrast, our method is orthogonal—compression is performed offline, and there is no additional inference-time overhead.
Other methods~\citep{Zhang2023H2OHO,atten_sink,li2024snapkvllmknowslooking} propose KV cache compression techniques that attend only to predefined token positions (e.g., start or end tokens). However, these strategies are incompatible with many-shot ICL, as different queries often rely on distinct and diverse demonstration sets. 
\vspace{-1.5mm}
\paragraph{Comparison with approaches to prompt compression for efficient inference.}
To address the computational challenges of long prompts, researchers have explored various prompt compression techniques.~\citet{chevalier2023adapting} introduced
AutoCompressors for recursively compressing long
text into summary vectors. However, their training involves adapting the main model, while we work in the setting where target-model is frozen.~\citet{ge2024context} proposed the In-context Autoencoder (ICAE), leveraging LLMs to encode contexts into memory slots. However, their work is based on a specific Prompt-with-Context (PwC) task they introduce in the paper, with contexts of length $< 1k$ tokens. Moreover, their aim is to provide a parameter efficient compressor, which we show in Section~\ref{sec:icae_comparison} is not effective for long-context many-shot compression.
\citet{jiang2023llmlingua} compress general prompts into concise natural language, whereas our aim is to provide compression for many-shot ICL task using memory-tokens, which don't necessarily translate into natural language.
\citet{mu2023learning} introduced "gisting," training LLMs to compress prompts into gist tokens. However, it requires modifying the main model, similar to~\citet{chevalier2023adapting}. Moreover, the evaluations are on very shot prompts of length $< 50$ tokens.
\citet{tan2024loco} proposed LLoCO, an offline context compression framework for long context, though trained particularly for task of document summarization. Moreover, their target-LLM requires LoRA updates, whereas we work in frozen-target LLM regime. 


While existing prompt compression methods have shown promise, they have not studied the task of long-context many-shot compression.
The complex interplay of numerous diverse examples within the many-shots require more fine-grained compression strategies with expressive compressors. Our proposed method, MemCom, aims to overcome these limitations by introducing a layer-wise compression approach coupled with a stronger compressor model.
We operate in the frozen-target LLM regime, which aligns with practical deployment scenarios such as on-device inference at the edge. In such cases, modifying the model weights is often undesirable, as it may interfere with the LLM's broader capabilities. Additionally, access to the original training data is rarely available, making any finetuning prone to overfitting and distributional drift. The most relevant prior work under this constraint is ICAE~\citep{ge2024context}, which we thoroughly compare with in Section~\ref{sec:experiments}.

\begin{table*}[t]
\centering
\caption{Datasets used in our work. We test each method on ICL tasks with varying large label sets and different domains. The average demonstration length is the average number of tokens using Mistral/Gemma tokenizer on each example shot.}
\resizebox{0.75\linewidth}{!}{
\begin{tabular}{ccccc}
\toprule
Dataset & Domain & \# Labels & Avg Demo Length & Example Output \\ \midrule
Trec-Coarse & Questions & 6 & 20.57/17.45 & abbreviation, entity\\
Trec-Fine & Questions & 47 & 20.38/18.11 & manner, reason\\
HWU64 & Conversational & 64 & 20.89/18.89 & music\_query, social\_post\\ 
Banking77 & Financial & 77 & 27.34/25.06 & change\_pin, request\_refund \\ 
Clinc150 & Multiple & 151 & 20.01/18.05 & cook\_time, income\\ \bottomrule
\end{tabular}
}
\label{tab:datasets}
\end{table*}
\section{Experimental Setup} \label{sec:exp_setup}
We conduct our experiments using two different LLM architectures: Gemma2-2B~\citep{gemmateam2024gemma2improvingopen} and Mistral-7B-v0.3~\citep{jiang2023mistral7b}, both using their base pre-trained checkpoints. To train our compressor models, we exclusively use a pretraining datasets, without any instruction tuning or task-specific finetuning. This stands in contrast to prior prompt compression works~\citep{ge2024context,tan2024loco,jiang2023llmlingua}, which typically relies on additional supervision and task specific finetuning to guide the compression process.
Our motivation for using pretraining data is threefold. First, in the context of ICL, labeled training data is often scarce or unavailable. Second, it is well-established that continued pretraining improves the downstream ICL capabilities of LLMs~\citep{brown2020language}. We view the compressor-assisted target model as an alternate architecture trained via next-token prediction: instead of attending to previous tokens explicitly, the model learns to predict next token by attending to the previous tokens' compressed representations. Thus, we hypothesize that training on large-scale pretraining data should also improve compression quality over time. Third, pretraining corpora are abundant, making it feasible to scale up training if resources permit—potentially improving compressor performance even more with further training.

The pretraining datasets used are a mix of FineWebEdu~\citep{finewebedu} and SlimPajama~\citep{slimpajama}. For downstream evaluations, we use a set of classification tasks with large label space. Table~\ref{tab:datasets} and Appendix~\ref{appnd:downstream_tasks} provide information of these tasks, including the number of classes and representative label examples. We do evaluations under two different sequence lengths: 3k tokens of many-shot input for Gemma2-2B and 6k tokens for Mistral-7B. To assess the effectiveness of compression, we experiment with three compression ratios: $3\times$, $6\times$, and $8\times$. Details of training hyperparameters such as learning rate, batch size, and weight decay are provided in Appendix~\ref{appnd:hyperparams}, and the training methodology for MemCom is described in Section~\ref{sec:memcom_arch}.

\section{MemCom Architecture and Training} \label{sec:memcom_arch}

MemCom consists of two LLM stacks forming a compressor, along with a set of learnable memory tokens. The first LLM, referred to as the Source-LLM, is initialized with copy of the target-LLM, and processes the many-shot input tokens to be compressed. The second LLM, the Memory-LLM, processes the memory tokens. It is initialized as a copy of the target-LLM as well, and in addition includes a randomly initialized 1-head cross-attention module at each layer. This module allows the memory tokens to attend to the corresponding layer representations from the Source-LLM. As a result, $m$ memory tokens extract compressed information from $t$ input tokens, producing a compact representation at each layer. The Target-LLM, which remains frozen during training and is used for inference, attends to these compressed memory tokens instead of the original many-shot input. This layer-wise transfer of compressed information enables the Target-LLM to leverage the many-shot context while avoiding the memory and compute overhead of attending to all $t$ input tokens explicitly.

The compression mechanism, depicted in Figure~\ref{fig:memcom} operates at each transformer layer. Specifically, for any given layer $i$, let $H_{source}^{i}\in\mathbb{R}^{t\times d}$ be the sequence of t source token input representations in the Source-LLM layer $i$. Similarly, let $H_{memory}^{i}\in\mathbb{R}^{m\times d}$ be the sequence of $m$ memory token representations after the Memory-LLM's self-attention module in layer $i$. The core of the compression is a cross-attention module where the memory tokens query the source tokens: $O^{i} = CrossAttention(Q=H_{memory}^{i}, K=H_{source}^{i}, V=H_{source}^{i})$. The Target-LLM then uses these layer-specific compressed representations $O_{i}$ as the key and value context for its attention modules, instead of attending to the original $t$ tokens.


We train MemCom using a standard next-token prediction loss on the target tokens. As mentioned in Section~\ref{sec:exp_setup}, only pretraining data is used for training. During training, for Gemma2-2B we sample 4k-token sequences and randomly select a split point within the range $[2.7\text{k}, 3.4\text{k}]$. The tokens before this point are assigned as source tokens and the remaining tokens as target tokens. For Mistral-7B, we sample 8k-token sequences and apply the same strategy, selecting a random split point in the range $[5.7\text{k}, 6.3\text{k}]$ for source/target token split. Training is conducted in two phases.

\noindent \textbf{Phase-1:} Since the cross-attention modules and memory tokens are randomly initialized—while the rest of the compressor is initialized from well-trained parameters, we begin by training only these randomly initialized components. This allows them to learn meaningful compression behavior without disrupting the pretrained parameters.

\noindent \textbf{Phase-2:} After initial convergence, we unfreeze the entire Source-LLM and Memory-LLM stacks and continue training both LLMs jointly, including the memory tokens and cross-attention layers.

Each phase is trained on approximately 80 billion tokens of pretraining data. The Target-LLM remains frozen throughout both phases. As we'll see in Section~\ref{sec:experiments}, most of the gains are achieved post Phase-1 training itself. While additional marginal gains can be achieved post Phase-2 training.

\begin{figure*}[th!]
   ~\centering
    \subfloat[]{\includegraphics[width=0.54\textwidth]{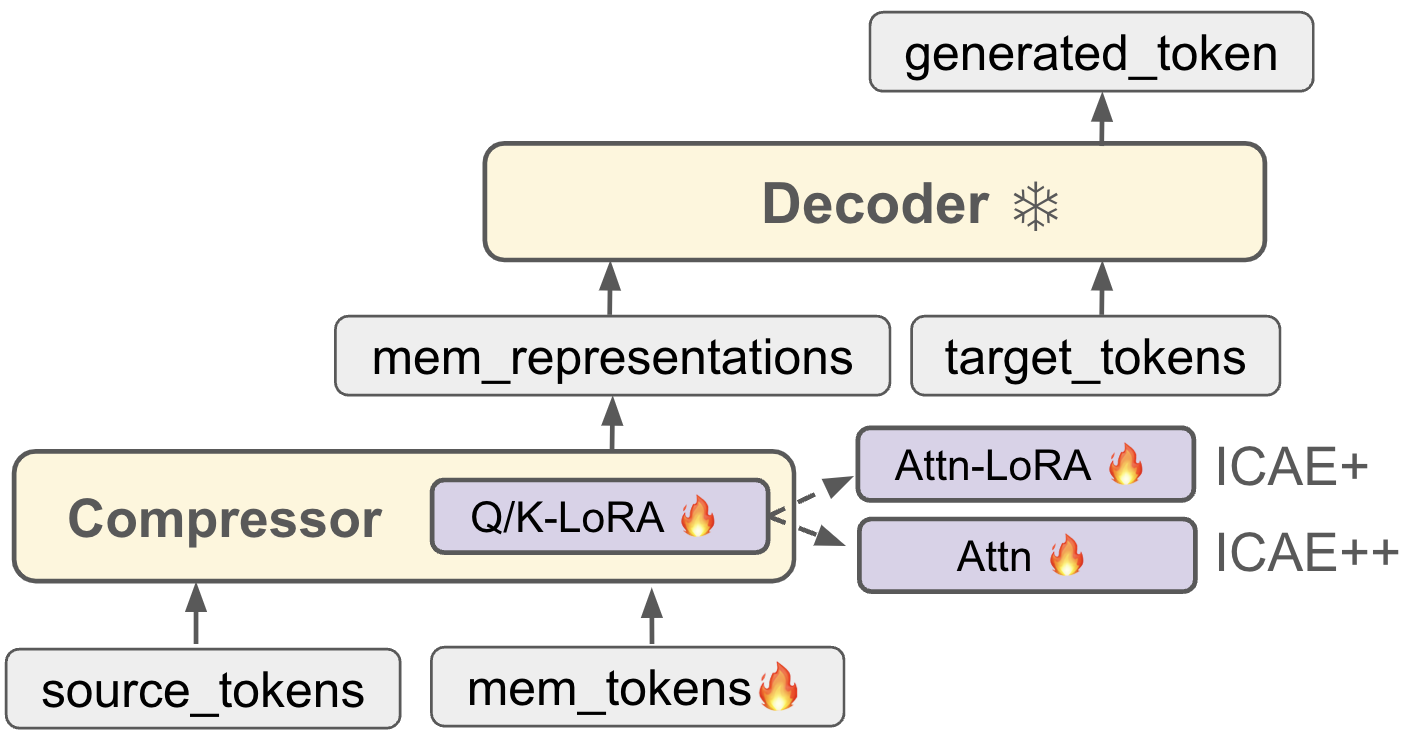} \label{fig:icae}}
    \hfill
    \subfloat[]{\includegraphics[width=0.44\textwidth]{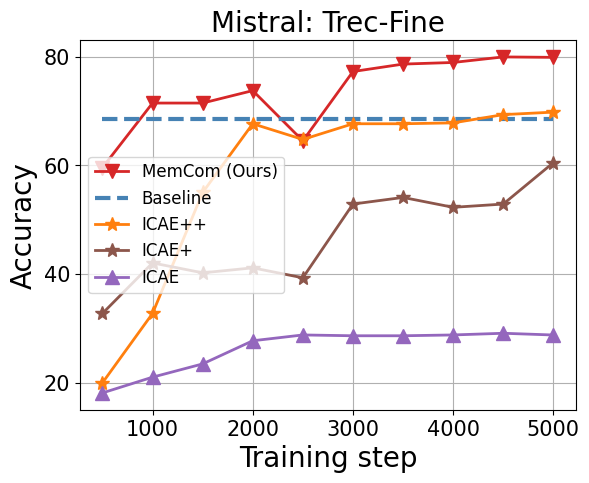} \label{fig:icae_vs_memcom_plot}}
   ~\caption{(a) Architecture of ICAE - The compressor is a copy of the decoder LLM, with LoRA parameters applied only to the Query and Key matrices. We extend this setup in ICAE+ by including LoRA parameters to value and post matrices as well. And we eventually extend to the entire attention module being trained, referred as ICAE++. (b) Performance improves as we move from \textit{ICAE} $\rightarrow$ \textit{ICAE+} $\mathbf{\rightarrow}$ \textit{ICAE++} $\rightarrow$ \textit{MemCom}. The comparison is shown through performance on Trec-Fine across training steps, on Mistral-7B while compressing 6k source tokens by $8\times$ into $768$ memory tokens.}
    \label{fig:icae_vs_memcom}
\end{figure*}

\section{Experiments} \label{sec:experiments}
\subsection{Baseline Study and Motivating Insights} \label{sec:icae_comparison}
We begin our analysis by evaluating ICAE~\citep{ge2024context} as a baseline for many-shot compression. ICAE introduces a compressor LLM, initialized with the same parameters as the target LLM but augmented with LoRA adapters on the query and key projection matrices (Figure~\ref{fig:icae}).
The input sequence is appended with $m$ learnable memory tokens, and a forward pass is performed through the compressor. The output memory token representations are treated as the compressed context. During inference, these $m$ compressed tokens are prepended to the target tokens in the frozen target LLM. Thereby, during inference the target tokens attend only to the compressed memory representations, rather than the full source sequence.

The ICAE training objective is similar to MemCom’s, using next-token prediction on the target tokens. In addition, ICAE includes an auto-encoding loss, where the model attempts to reconstruct the original source tokens from the compressed memory representation. For a detailed description of ICAE's architecture and training procedure, we refer the reader to \citet{ge2024context}. In our work, we train ICAE with just next-token prediction loss. Using auto-encoder loss in addition led to instability at higher learning rate, and non-optimal training at lower learning rate (Appendix~\ref{appnd:hyperparams}). 

To understand the limitations of ICAE clearly, we study it at the larger scale of 7B Mistral model, longer sequence length of compressing $6k$ source tokens, and large compression ratio of $8\times$. Training follows the original setup in \citet{ge2024context} (except using only next token prediction loss), keeping LoRA rank as 32, and using similar training as our MemCom training procedure. Under the more challenging many-shot setting, we find that ICAE fails to perform meaningful compression, performing significantly lower than a simple baseline that fits fewer full shots within the same token budget. This suggests that the ICAE compressor, as originally proposed, is not powerful enough for compressing long many-shot sequences.


To verify the above hypothesis, we extend the LoRA parameters to value and post matrices as well (in addition to just query and key matrices as in original ICAE setup) in the attention module. We refer to this architecture as ICAE+. As seen in Figure~\ref{fig:icae_vs_memcom_plot} and Appendix~\ref{appnd:icae_comparison}, the performance increases, confirming our hypothesis that long-context many-shot compression requires a stronger compressor. To test this even further, we make the entire attention module trainable in the compressor network of ICAE, rather than just adapting it via LoRA. We notice a significant performance boost as a result of this, with performance starting to reach close to the simple baseline. We call this modified ICAE architecture as ICAE++.

Next, we motivate our design choice of layer-wise compression, as opposed to using the final-layer output of a full forward pass as the compressed representation. In a vanilla ICL setting,
the source tokens would be prepended to the target tokens, and the model would process the combined sequence through all layers. This means that, at each layer, the target tokens have access to rich and evolving representations of the source tokens.

In the compressed setting, however, the source tokens are removed and replaced with a compact representation. To preserve performance, the ideal scenario would be for the target tokens to receive representations that closely approximate the original source-token representations at each layer.
\begin{table*}[!t]
\centering
\caption{MemCom performance on Mistral-7B across different compression ratios. The many-shots sequence length to compress is fixed at $6k$ tokens. The second column refers to the number of tokens $m$ being attended to in each layer by the target tokens. For baseline, this means using many-shots comprising $m$ tokens. For ICAE++ and MemCom, it refers to $m$ memory tokens. $m=2k, \; 1k, \; 768$ tokens represent $3\times,\; 6\times,\text{ and } 8\times$ compression ratio respectively.}
\resizebox{0.75\textwidth}{!}{
\begin{tabular}{ccccccc}
\toprule
Task $\rightarrow$ & $m\downarrow$ & {\small TREC-Coarse} & {\small TREC-Fine} & {\small HWU64} & {\small Banking77} & {\small Clinc-150} \\ \midrule
Baseline & $6k$ & 86.58 & 82.71 & 88.36 & 83.53 & 87.05\\ \midrule
Baseline & \multirow{4}{*}{$2k$} & 83.53 & 75.89 & 84.32 & 74.81 & 78.34\\
ICAE++ & & 80.56 & 75.31 & 85.53 & 75.71 & 82.55 \\
\textbf{MemCom} & & 83.33 & 80.03 & 85.27 & 79.61 & \textbf{84.37} \\
\textbf{MemCom-P2} & & \textbf{85.01} & \textbf{80.64} & \textbf{86.56} & \textbf{80.27} & 83.72\\ \midrule
Baseline & \multirow{4}{*}{$1k$} & 82.20 & 70.12 & 80.16 & 59.22 & 72.75\\
ICAE++ & & 78.55 & 72.40 & 83.01 & 71.51 & 77.15\\
\textbf{MemCom} & & 84.50 & \textbf{79.12} & \textbf{83.58} & 75.32 & 81.06 \\
\textbf{MemCom-P2} & & \textbf{86.20} & 77.01 & 83.12 & \textbf{75.39} & \textbf{81.95}\\ \midrule
Baseline & \multirow{4}{*}{$786$} & 82.88 & 68.60 & 72.50 & 54.09 & 69.81\\
ICAE++ & & 78.01 & 69.70 & 83.18 & 69.71 & 77.25\\
\textbf{MemCom} & & \textbf{83.31} & 79.27 & 82.97 & 74.35 & 81.32 \\
\textbf{MemCom-P2} & & 83.30 & \textbf{79.80} & \textbf{83.89} & \textbf{75.42} & \textbf{82.38}\\ \bottomrule
\end{tabular}
}

\label{tab:mistral_results}
\end{table*}

This motivates our approach in MemCom, where we perform layer-wise compression. Specifically, we introduce a single-head cross-attention module in each layer of the compressor. This module allows the memory tokens to attend to the Source-LLM’s representations at the corresponding layer and produce compressed outputs tailored for that layer in the Target-LLM. This fine-grained, layer-wise transfer of information enables the model to more faithfully approximate the behavior of the original many-shot attention. Note that inference-time latency of our layer-wise compression and ICAE's full forward pass compression remains the same - as the target tokens attend to $m$ tokens in each layer in both the algorithms.
Figure~\ref{fig:icae_vs_memcom_plot} demonstrates that transitioning from the ICAE++ architecture to our MemCom approach with layer-wise cross-attention, trained under the same Phase-1-only setting, leads to a clear performance improvement. This result supports our central hypothesis that fine-grained, layer-wise compression is substantially more effective for many-shot ICL. To further validate this hypothesis, we present additional experiments in Section~\ref{sec:evals}.

\subsection{Evaluations} \label{sec:evals}
For evaluation, we compare our method against two baselines: (1) the vanilla baseline of fitting fewer full shots within the compressed token budget, and (2) ICAE++, our stronger variant of ICAE. As a reference point, we also report the performance when using all the source tokens without any compression, which serves as a natural upper bound for the task.

We present results for Mistral-7B in Table~\ref{tab:mistral_results} and Gemma2-2B in Table~\ref{tab:gemma_results} . For the Clinc-150 dataset, we found that it was not possible to fit at least one example per class within the 3k-token source budget used for Gemma2-2B. Since the full source sequence itself would be incomplete, evaluating compression methods under this setting would not be meaningful. Therefore, we report Clinc-150 results only for Mistral-7B, where the 6k-token source budget is sufficient to include at least one example from all classes.

In both Table~\ref{tab:mistral_results} and Table~\ref{tab:gemma_results}, the second column indicates the number of tokens $m$ that the target tokens attend to at each layer. For the baseline, this corresponds to fitting as many full-shot examples as possible within $m$ tokens. For ICAE++ and MemCom, it represents the number of memory tokens. MemCom-P2 refers to MemCom after Phase-2 training. We observe the following resuts:

\begin{table*}[!t]
\centering
\caption{MemCom performance on Gemma2-2B across different compression ratios. The many-shots sequence length to compress is fixed at $3k$ tokens. $m=1k, \; 512, \; 384$ tokens represent $3\times,\; 6\times,\text{ and } 8\times$ compression ratio respectively. MemCom-P2 refers to MemCom after Phase-2 training.}
\resizebox{0.7\textwidth}{!}{
\begin{tabular}{cccccc}
\toprule
Task $\rightarrow$ & $m\downarrow$ & TREC-Coarse & TREC-Fine & HWU64 & Banking77 \\ \midrule
Baseline & $3k$ & 88.31 & 75.77 & 77.72 & 41.70\\ \midrule
Baseline & \multirow{4}{*}{$1k$} & 83.24 & 70.13 & \textbf{65.02} & 35.63\\
ICAE++ & & 76.11 & 66.16 & 61.8 & 33.52\\
\textbf{MemCom} & & 82.01 & 72.12 & 46.34 & 30.20\\
\textbf{MemCom-P2} & & \textbf{87.78} & \textbf{74.24} & 53.75 & \textbf{38.05}\\ \midrule
Baseline & \multirow{4}{*}{$512$} & 60.83 & 39.35 & 49.06 & 31.04\\
ICAE++ & & 72.22 & 55.95 & \textbf{52.81} & 31.49 \\
\textbf{MemCom} & & 76.22 & 70.80 & 47.61 & 41.69 \\
\textbf{MemCom-P2} & & \textbf{80.21} &\textbf{72.09} & 49.28 & \textbf{47.63} \\ \midrule
Baseline & \multirow{4}{*}{384} & 16.11 & 2.74 & 1.64 & 1.36\\
ICAE++ & & 68.93 & 56.71 & \textbf{53.05} & 29.94 \\
\textbf{MemCom} & & \textbf{75.60} & 68.34 & 35.31 & 39.55 \\
\textbf{MemCom-P2} & & \textbf{75.60} & \textbf{68.91} & {36.17} & \textbf{42.43}\\ \bottomrule
\end{tabular}
}
\label{tab:gemma_results}
\end{table*}

\textbf{First}, we observe that moving from compression through a full-model forward pass, as in ICAE++, to layer-wise compression, as in MemCom after Phase-1 training, consistently leads to improved performance. Specifically, MemCom outperforms ICAE++ across all benchmarks on Mistral-7B when compressing 6k source tokens. On Gemma2-2B, the same trend holds across all tasks except HWU64. Investigating this reverse trend on HWU64 remains an open direction for future work.

\textbf{Second}, we observe that MemCom Phase-2 training generally leads to improved performance over Phase-1 training. This effect is more pronounced at lower compression ratios on Gemma2-2B. However, as we move to larger source sequence lengths (6k tokens) and stronger models like Mistral-7B, we find that Phase-1 training already provides the majority of the gains, with Phase-2 offering marginal additional improvements. This is encouraging, as it suggests that a lightweight compressor—comprising a simple 1-layer cross-attention module per layer—is sufficient to capture most of the benefits of many-shot compression at scale.

\textbf{Third}, we find that while the performance of the baseline and ICAE++ degrades sharply with increasing compression ratios, MemCom maintains high accuracy with minimal performance drop (Figure~\ref{fig:compression_fig}). This effect is particularly pronounced with stronger models like Mistral-7B and longer source sequences of 6k tokens. Notably, this robustness at larger sequence lengths highlights the practical utility of MemCom for real-world long-context ICL applications.



\paragraph{Ablation: Cross-attention module design choice.} In Appendix~\ref{appnd:ablation}, we study the effect of varying the structure of the cross-attention module used in MemCom to mult-head attention (MHA) or multi-query attention (MQA). Our conclusion is using 1-head cross-attention provides the best performance overall, and we therefore adopt this configuration for all other experiments in the paper.

\section{Limitations and Future Work} \label{sec:future_work}
We primarily evaluate MemCom on classification tasks with large output spaces (Table~\ref{tab:datasets}). While we conducted preliminary experiments on several generation tasks—including summarization (XSum~\citep{xsum}), question answering (TriviaQA~\citep{triviaqa}), sequential reasoning (Sequential Parity~\citep{agarwal2024many}), and arithmetic reasoning (GSM8K~\citep{gsm8k})—we observed limited differences between the baselines. Specifically, the performance of a baseline using fewer shots (to fit within the compressed token budget) was already comparable to a baseline using all shots without compression. In such cases, where the uncompressed baseline itself shows minimal sensitivity to the number of shots, it is unsurprising that compression does not yield substantial gains. We believe that extending our approach to even longer sequence lengths, beyond those explored here, could make many-shot compression more impactful for generation tasks as well. This is since the performance gap between using fewer and more shots is likely to widen with larger contexts.

Another limitation is MemCom's substantial training cost (80-160B tokens), necessary for building powerful compressors. This cost will escalate with longer sequences due to increased token processing and quadratic attention complexity. Thus, an important direction for future work is to explore strategies for making compressor training more efficient.

Third, many-shot compression may be unsuitable for tasks that require precise access to individual input tokens. For example, in low-resource language to English translation tasks~\citep{agarwal2024many}, ICL operates by mapping specific tokens from the source language to their English counterparts. As the model needs to preserve fine-grained token information rather than relying on a compressed abstraction, in our preliminary experiments we noticed degraded performance on such tasks.


\section{Conclusion}
In this work, we studied the problem of compressing the information contained in many-shot prompts for In-Context Learning (ICL). We showed that existing prompt compression methods, when applied directly, fail to perform effectively in the many-shot setting. To address this, we proposed MemCom, a novel approach that employs a more powerful compressor and performs layer-wise compression to produce fine-grained representations. In contrast to prior methods that use the final embedding of a compressor model as a static summary, MemCom transfers compressed information at each layer, enabling more faithful reconstruction of the original context.

Through extensive experiments on classification tasks with large label spaces, we demonstrated that MemCom consistently outperforms strong baselines. Notably, at high 8x compression ratios where baseline accuracy can plummet by over 70\%, MemCom maintains robust performance with accuracy drops typically below 10\%.
This provides a promising path toward efficient utilization of many-shot ICL, without incurring the substantial memory and compute costs of the vanilla setting.

\section{Acknowledgment}
We would like to thank Peyman Milanfar, Arun Suggala, and Sanjiv Kumar for their valuable feedback on an earlier version of this paper.

\bibliography{custom}
\newpage
\appendix

\begin{figure*}[t]
    \centering
    \subfloat[ICAE++ training curve under different learning rates, when using both next token prediction loss and autoencoder loss during training. We see that the largest learning rate having stable training is $5e-6$.]{\includegraphics[width=0.47\textwidth]{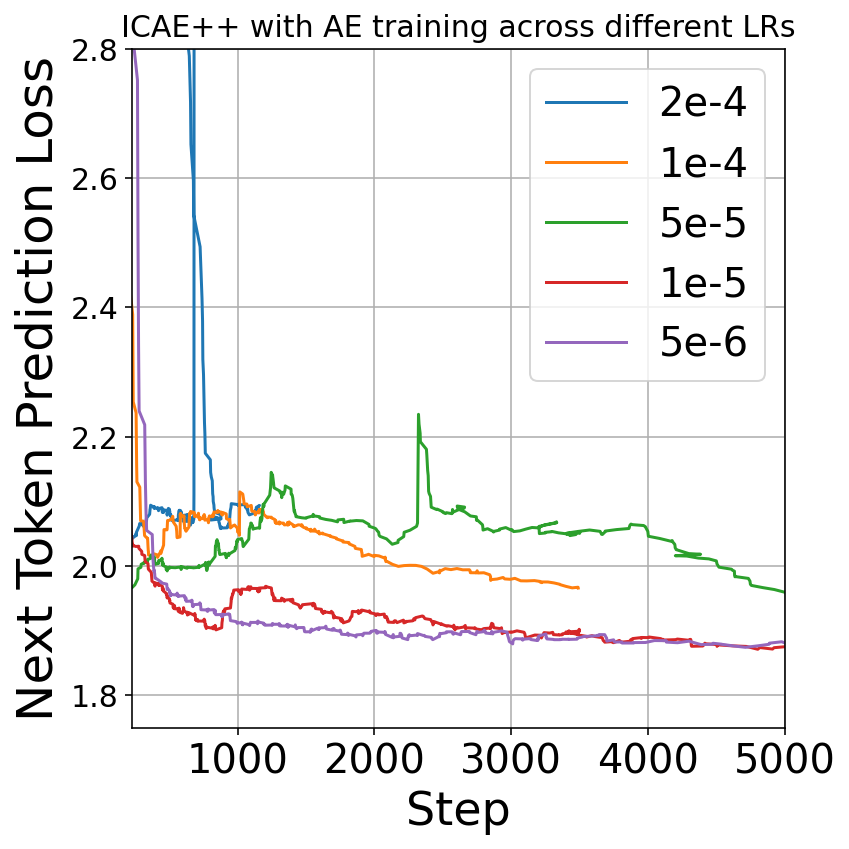} \label{fig:icae_pp_train_losses}}
    \hfill
    \subfloat[Comparison of compression performance across different compression ratios on Mistral model for Trec-Coarse benchmark. We show the remaining plots in Figure~\ref{fig:compression_fig} ]{\includegraphics[width=0.5\textwidth]{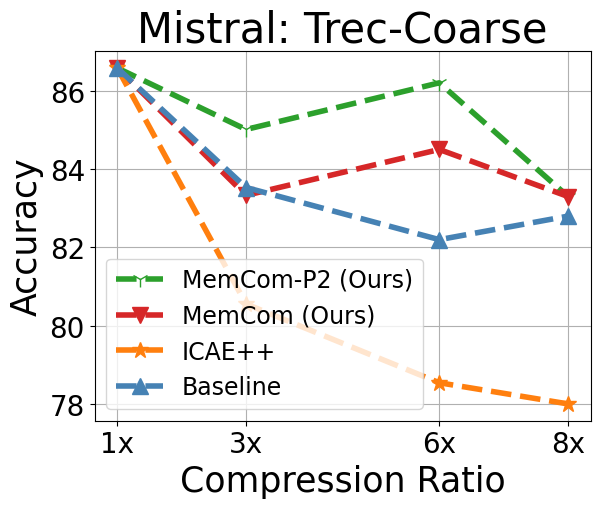}}
    \caption{}
\end{figure*}
\section{Experimental Setup}
\subsection{Training}
For training MemCom and ICAE++, we use a mixture of FineWebEdu~\citep{finewebedu} and SlimPajama~\citep{slimpajama} datasets. For Mistral training, we filter all sequences that are smaller than 7k tokens, and trim them to 8k sequence lengths (using padding if needed). During training, we randomly split each sequence between $[5.7k, 6.3k]$ tokens as source tokens, and the remaining as target tokens. We use $m$ memory tokens, where $m\in\{2048, 1024, 768\}$ for Mistral-7B.

Similarly, for Gemma-2 2B we filter out all sequences less than 3.7k, and trim them to 4k sequence lengths. Source tokens are split in range $[2.7k, 3.4k]$ and the remaining are considered as target tokens. Number of memory tokens $m \in \{1024, 512, 384\}$

We used 512 TPUv5 accelerator~\citep{GoogleTPUv5eBlog2023} for all of our training runs. Phase-1 of MemCom training and ICAE++ took around 12/18 hours for Gemma/Mistral model, whereas Phase-1 of MemCom took 24/36 hours.

\subsection{Hyperparameters} \label{appnd:hyperparams}
Batch size for Mistral is kept at $1024$, and for Gemma at $2048$. Weight decay is kept at 0 for all experiments. Learning rate (LR) is tuned in the range $\{2e-4,\; 1e-4,\; 5e-5,\; 2e-5,\; 1e-5,\; 5e-6,\; 1e-6,\; 8e-7\}$. We use the largest learning rate possible that does not lead to instability. For all the MemCom Phase-1 training, this happens to be $2e-4$. For MemCom Phase-2 training, this was $2e-6$, except $8\times$ compression ratio on Mistral where this was $8e-7$. For ICAE++, as mentioned in Section~\ref{sec:icae_comparison}, training with Autoencoder loss led to unstable training at higher learning rate, and was relatively stable at $5e-6$. We show the next token prediction train loss plots across training steps in Figure~\ref{fig:icae_pp_train_losses} for ICAE++ when AE loss was enabled as well during training. Without AE loss, and only using next token prediction loss, ICAE++ training was stable with LR = $2e-4$. A warmup of 1.5k steps was used for ICAE++,  0.5k for MemCom Phase-1, and 1.5k for MemCom Phase-2.

\subsection{Downstream Tasks} \label{appnd:downstream_tasks}
We use 5 downstream classification tasks with large label space. These are: TREC-Coarse and TREC-fine~\citep{li-roth-2002-learning,hovy-etal-2001-toward}, HWU64~\citep{hwu64}, Banking77~\citep{banking77}, Clinc150~\citep{clinc150}. More details are provided in Table~\ref{tab:datasets}.

For each dataset, the construction of few-shot input prompts, constrained to $t$ tokens, involved a round-robin sampling strategy ensuring class balance. We iteratively selected one random shot per class. This process was repeated until the token budget was nearly filled. In the concluding iteration, to precisely meet the $t$-token limit, if a selected shot exceeded the budget, it was dropped and the loop to add shots was ended. Thus, we few-shot prompt construction followed a class-balanced sampling procedure. 

We first created a unified candidate pool by merging its original train and test splits. From this pool, 20 distinct instances per class were randomly selected to serve as our evaluation queries. For each selected query, a many-shot demonstration prompt of up to t tokens was constructed using the method detailed above.

\section{Dataset and Model Licenses}
All the datasets and models used are publicly available and were utilized in accordance with their respective licenses, all of which permit research use. Specifically, Banking77, TREC, and HWU64 is available under the Creative Commons Attribution 4.0 International (CC BY 4.0) license. The Clinc150 dataset is available under Creative Commons Attribution 3.0 International (CC BY 3.0) license. FineWebEdu is under Open Data Commons Attribution License (ODC-By) v1.0 license and under CommonCrawl's Terms of Use, and SlimPajama is under Apache 2.0 license.

Mistral has Apache license 2.0 and Gemma has Gemma Terms of Use as its license, both permitting use of the models for research purposes.

\section{Baseline Study and Motivating Insights} \label{appnd:icae_comparison}
As mentioned in Section~\ref{sec:icae_comparison}, we study ICAE and gradually change its architecture to enable more powerful compressor. Due to resource constraints, we do our study on just the most relevant setting - using the larger model Mistral-7B, longest source length sequence of $6k$ tokens, and highest compression ratio of $8\times$.
As we see in Figure~\ref{fig:icae_vs_memcom_plot}, moving from \textit{ICAE} $\rightarrow$ \textit{ICAE+} $\mathbf{\rightarrow}$ \textit{ICAE++} $\rightarrow$ \textit{MemCom} results in improved performance. We also show this improvement across other benchmarks in Table~\ref{tab:icae_to_icae_pp}.

Moreover, as mentioned in Section~\ref{appnd:hyperparams}, ICAE++ with Autoencoder loss performs worse compared to without Autoencoder loss. We show this performance difference in Table~\ref{tab:icae_pp_orig_vs_icae_pp}.

\begin{table*}[t]
\begin{tabular}{|w{c}{2.5cm}|w{c}{2cm}|w{c}{1.6cm}|w{c}{1.4cm}|w{c}{1.4cm}|w{c}{1.4cm}|}
\toprule
Task $\rightarrow$ & {\small TREC-Coarse} & {\small TREC-Fine} & {\small HWU64} & {\small Banking77} & {\small Clinc-150} \\ \midrule
Baseline-$6k$ & 86.58 & 82.71 & 88.36 & 83.53 & 87.05\\ \midrule
Baseline-$768$ & 82.88 & 68.60 & 72.50 & 54.09 & 69.81\\
ICAE & 51.67 & 28.81 & 32.34 & 20.97 & 34.46 \\
ICAE+ & 71.11 & 56.10 & 54.43 & 34.03 & 44.74 \\
ICAE++ & 78.01 & 69.70 & 83.18 & 69.71 & 77.25\\
MemCom & 83.31 & 79.27 & 82.97 & 74.35 & 81.32\\\bottomrule
\end{tabular}
\centering
\caption{
Baseline-$6k$ refers to the baseline using all 6k source tokens worth of shots. Baseline-$768$ is the $8\times$ compression comparable baseline, which uses $8\times$ less tokens worth of shots. Performance increases as we move from \textit{ICAE} $\rightarrow$ \textit{ICAE+} $\mathbf{\rightarrow}$ \textit{ICAE++} $\rightarrow$ \textit{MemCom}. This is because of having a strong model (\textit{ICAE} $\rightarrow$ \textit{ICAE+} $\mathbf{\rightarrow}$ \textit{ICAE++}) and more fine grained information transfer through layer-wise compression (\textit{ICAE++} $\rightarrow$ \textit{MemCom}).
}
\label{tab:icae_to_icae_pp}
\end{table*}

\begin{table*}[t]
\begin{tabular}{|w{c}{2.7cm}|w{c}{2cm}|w{c}{1.6cm}|w{c}{1.4cm}|w{c}{1.4cm}|w{c}{1.4cm}|}
\toprule
Task $\rightarrow$ & {\small TREC-Coarse} & {\small TREC-Fine} & {\small HWU64} & {\small Banking77} & {\small Clinc-150} \\ \midrule
Baseline-$6k$ & 86.58 & 82.71 & 88.36 & 83.53 & 87.05\\ \midrule
Baseline-$768$ & 82.88 & 68.60 & 72.50 & 54.09 & 69.81\\
ICAE++ with AE & 58.33 & 32.62 & 17.73 & 22.60 & 27.75 \\
ICAE++ & 78.01 & 69.70 & 83.18 & 69.71 & 77.25\\ \bottomrule
\end{tabular}
\centering
\caption{
Baseline-$6k$ refers to the baseline using all 6k source tokens worth of shots. Baseline-$768$ is the $8\times$ compression comparable baseline, which uses $8\times$ less tokens worth of shots. 
ICAE++ with AE refers to the ICAE++ architecture trained with Autoencoder loss in addition to next token prediction loss. We observe that ICAE++ performs better without Autoencoder loss.
}
\label{tab:icae_pp_orig_vs_icae_pp}
\end{table*}

\section{Ablation} \label{appnd:ablation}

\begin{table*}[ht!]
\centering
\caption{MemCom comparison on Mistral-7B with different cross-attention modules, on $8x$ compression ratio. Each experiments is trained on Phase-1 training only. The Baseline performance in the second row is using all $6k$ source tokens. MQA* refers to MQA initialized with same parameters as the model's self-attention module.
}
\begin{tabular}{cccccc}
\toprule
Task $\rightarrow$ & {\small TREC-Coarse} & {\small TREC-Fine} & {\small HWU64} & {\small Banking77} & {\small Clinc-150} \\ \midrule
Baseline & 86.58 & 82.71 & 88.36 & 83.53 & 87.05\\ \midrule
1-head attention & \textbf{83.31} & \textbf{79.27} & 82.97 & 74.35 & 81.32 \\
MHA & 75.11 & 69.36 & \textbf{85.23} & 74.82 & 83.41 \\
MQA & 72.23 & 62.04 & 83.64 & \textbf{75.36} & 83.22\\
MQA* & 77.78 & 69.82 & 71.41 & 62.99 & \textbf{83.38}\\ \bottomrule
\end{tabular}

\label{tab:cross_attention_ablation}
\end{table*}

\paragraph{Cross-attention module design choice.} We study the effect of varying the structure of the cross-attention module used in MemCom. Specifically, we compare using multi-head attention (MHA), multi-query attention (MQA), and our default 1-head attention, with all variants trained from scratch. Since Mistral employs MQA in its self-attention layers, another plausible design choice is to initialize the MQA cross-attention with the same parameters as the model's MQA self-attention. However, we find that this variant also doesn't outperforms 1-head cross-attention overall.

Based on the results shown in Table~\ref{tab:cross_attention_ablation}, we conclude that using 1-head cross-attention, initialized from scratch, provides the best performance overall. We therefore adopt this configuration for all other experiments in the main paper.

\end{document}